\documentclass{elsart}

\usepackage{amsmath}
\usepackage{amssymb}
\usepackage{dsfont}
\usepackage{cite}%
\usepackage{array}
\usepackage{graphicx}
\usepackage{enumerate}
\theoremstyle{plain} 

\theoremstyle{plain} 

\theoremstyle{plain} 

\theoremstyle{plain} 

\theoremstyle{definition} 

\theoremstyle{definition} 
\journal{AMC}
\begin{document}

\begin{frontmatter}

\title{\textbf{On the performance of a hybrid genetic algorithm in dynamic environments}}
\author{Quan Yuan\corauthref{cor}},\,\ead{quanyuan@wayne.edu}
\author{Zhixin Yang}\ead{zhixin.yang@wayne.edu}

\corauth[cor]{Corresponding author. {\em Tel.: +1-313-577-3300}}
\address{Department of Mathematics, Wayne State University,\\ Detroit, Michigan 48202, USA}

\begin{abstract}
The ability to track the optimum of dynamic environments is
important in many practical applications. In this paper, the
capability of a hybrid genetic algorithm (HGA) to track the optimum
in some dynamic environments is investigated for different
functional dimensions, update frequencies, and displacement
strengths in different types of dynamic environments. Experimental
results are reported by using the HGA and some other existing
evolutionary algorithms in the literature. The results show that the
HGA has better capability to track the dynamic optimum than some
other existing algorithms.
\end{abstract}
\begin{keyword}
Hybrid genetic algorithm (HGA), dynamic environments, optimization
\end{keyword}
\end{frontmatter}
\section{Introduction}

Traditionally, research on optimization problems has been focused on
stationary cases, i.e., problems are precisely given in advance and
remain fixed during the optimization processes. However, in many
practical optimization problems, a wide range of uncertainties
should be taken into account
(\cite{branke2001,morrison2004,chan2011148,chan2012}). For example,
the fitness evaluation is subject to noise, the design variables are
subject to perturbations or changes after the optimal solution has
been determined (e.g. due to manufacturing tolerances). The
environmental conditions may change over time due to some factors
such as the stochastic arrival of new tasks, machine faults,
climatic changes, or financial issues. These problems are called
dynamic optimization problems.

In this paper, we focus on a situation that the fitness function is
deterministic at any point in time, but is dependent on time $t$.
Usually, for stationary optimization problems, the aim is to design
an optimization algorithm that can quickly and precisely locate the
optimum solution(s) in the search space. However, for dynamic
optimization problems, the situation is quite different. Since the
optimum may change over time, this kind of optimization problems has
{\em dynamically changing optima}. An optimization algorithm used in
this dynamic environments should be able to adapt itself during
optimization. That is, it should not only find the optima but also
track their trace in the solution space as closely as possible
rather than requiring a repeated restart of the optimization
process.

A genetic algorithm (GA) is a generic population-based
meta-heuristic optimization algorithm. Candidate solutions to the
optimization problem play the role of individuals (parents) in a
population. Some mechanisms inspired by biological evolution:
selection, crossover and mutation are used. The fitness function
determines the environment within which the solutions ``survive''.
Then new groups of the population (children) are generated after the
repeated application of the above operators. GAs have attracted
great interests in the past decades. They are often used for solving
stationary optimization problems and there are also many successful
applications. Since GAs only use fitness as an indicator to operate
individuals in population, they have implicit parallelism,
robustness, and better ability of global searching than other
heuristics \cite{koza1990}. Moreover, GAs are population-based and
stochastic algorithms. So they have adaptivity in some way and
researchers think they can be applied on dynamic optimization
problems after some modification.

Solving dynamic optimization problems using GAs has received
increasing interest. Some methods have been proposed in recent
years. These methods can be roughly grouped into three classes:
\begin{enumerate}[1.]
\item Modify or adapt some parameters of GAs\cite{cobb1993523,grefenstette19992031,krishnakumar1989289}.
\item Memorize the past using information of GAs (\cite{goldberg198759,dasgupta1992145,mori1999240,branke19991875}),
explicit/implicit memory
(\cite{smith1987175,ng1995159,wineberg20003}), or multiple
populations (\cite{ursem200019,branke2000299}).
\item Maintain population diversity by, such as, inserting randomly generated individuals \cite{grefenstette1992137}, niching \cite{cedeno1997361},
or reformulating the fitness function considering the age of
individuals \cite{ghosh1998666} or the entropy of the population
\cite{mori1998149}.
\end{enumerate}

For more information, we refer to \cite{jin2005303},
\cite{woldesenbet2009500}, and the references there in.

Hybrid genetic algorithms (HGAs) belong to a type of GA in which a
local search is embedded as an auxiliary. The genetic search is in
charge of the broad search, while the local search is referred to as
the deep search. The results obtained by the HGA are often better
than those gained by a GA without a local search \cite{gen2000}. It
is naturally thought that a GA used in dynamic environments can be
embedded a local search to improve its performance. And the local
search can also help the algorithm track the trace of optima to a
certain extent. However, to the best of the authors' knowledge,
using HGA in dynamic environments is studied by a few researchers.
Garrett and Walker's work \cite{garrett2002359} is the only one the
authors can find.

Recently, a new kind of HGA, which can solve a class of stationary
optimization problems, has been proposed by Yuan {\it et al.}
(2008)\cite{yuan2008924}. A new strategy which has adaptive ability
is applied in this HGA. In this paper, we will investigate the
algorithm's performance when it is applied in dynamic environments
and compare its performance with some other existing algorithms. The
HGA is described in Section 2. New dynamic environments are
presented in Section 3. The results and conclusions are presented in
Section 4 and Section 5, respectively.

\section{The hybrid genetic algorithm}
In this section, we will describe the HGA in \cite{yuan2008924}.
Consider the following optimization problem:
\begin{equation}\label{eq1}
\min\:f(\boldsymbol{x})\;\;s.t.\,\boldsymbol{x}\in S
\end{equation}
where $S$ is an $n$-dimensional box constraint, i.e., $S = \{
\boldsymbol{x} = (x_i )_{n \times 1}  \in \mathds{R}^n |a_i  \le x_i
\le b_i ,i = 1, \cdots ,n\, \}$, and $f(\boldsymbol{x})$ is a
continuous and multimodal function on $S$.

Let
\[
x_i  = \frac{{(a_i  + b_i )}}{2} + \frac{{(a_i  - b_i )}}{2}\sin y_i
\,,\,y_i  \in \mathds{R}\,,\,i=1,\ldots,n.
\]
Then problem \eqref{eq1} can be transformed into an unconstrained
optimization problem
\begin{equation}
\min\:f(\boldsymbol{y})\;\;s.t.\,\boldsymbol{y}\in \mathds{R}^n
\end{equation}

For the HGA used in this study, an individual $X_i$ is an object
variable $\boldsymbol{y}$.

The following steps of the HGA are proposed:

\begin{enumerate}[\rm\bf {Step} 1]
\item(Initialization) {\rm Define the {\em population size} $N$ and the {\em maximal generation number}; randomly generate $N$ feasible \textit{individuals}
as the {\em initial population} $\overrightarrow X
(0)=\{X_1,X_2,\cdots,X_N\} $; set the initial generation number
$t_{gen}=0$.}
\item(Local-search) {\rm Perform a local search for every individual $X_i$ in $\overrightarrow X (t_{gen})$ to obtain the local optimum $X_i^{\#}$.}
\item(Evaluation) {\rm Evaluate $X_i$ in $\overrightarrow X (t_{gen})$ by
$f(X_i^{\#})$}, $X_i$ itself is not changed.
\item\mbox{}\\
\vspace{-1cm}
\begin{enumerate}
\item{Selection}{\rm: Choose $N$ pairs of parents.}
\item{Reproduction (Crossover or Mutation)}{\rm: for pairs of parents $Y_1$ and
$Y_2$, who have the same fitness value and have some local optimum
$X_i^{\#}$, a mutation step is performed to generate a child;
otherwise, a crossover step is performed.

Selection and reproduction will generate a new population.}
\end{enumerate}
\item {\rm Stop the algorithm if the maximal generation number is reached,
or let $t_{gen}=t_{gen}+1$, and return to Step 2.}
\end{enumerate}

We use a traditional optimization method, such as the The
Broyden-Fletcher-Goldfarb-Shanno (BFGS) algorithm, in Step 2. {\em
Tournament selection} is performed in Step 4(a). In Step 4(b), the
method of crossing two parents $Y_1$ and $Y_2$ is as below:
\begin{equation}
Y'=\alpha Y_1+(1-\alpha)Y_2
\end{equation}
where $\alpha$ is a random number satisfying $0\le\alpha\le1$.

The mutation step should be designed in order to let the individuals
have the chance to be better. But excessive mutation will destruct
some good individuals. So in the HGA, the mutation step is executed
as
\begin{equation}\label{mutation}
Y'=Y+\pi\cdot(Y_1-Y_2)
\end{equation}
where $Y$ is either one of the two parents. If the norm
$\|Y_1-Y_2\|$, where $\|\cdot\|$\\ denotes the Euclidian norm, is
too small, a random $\boldsymbol{d}=(d_1,\ldots,d_n)$ is generated
where each $d_i$ is taken $1$ or $-1$. We take this mutation step
because in the later stage of the HGA, some individuals will tend to
converge. And the norm $\|Y_1-Y_2\|$ in \eqref{mutation} will
decrease. At this situation, the mutation step will be less
destructive to some good individuals.

The mechanism of the crossover and mutation operations are based on
avoiding ``premature''. Yuan et al. in \cite{yuan2008924} tried to
let the population scatter on the feasible set. They considered two
individuals are ``similar'' if a local search can find the same
local optimum starting from both of them; in another words, they
belong to the same local optimum¡¯s ``neighborhood''. If two parents
selected by a GA are similar, the mutation step is executed to
generate a new child. The GA tries to explore a new area of the
feasible domain. Otherwise, if two parents are not similar, a
crossover step is performed. The crossover and mutation steps may
find a better region that includes a better local optimum, and which
can be found by a local search. A further detailed discussion of the
HGA in \cite{yuan2008924} can be found in \cite{yuan2010640}.

The following aspects are the motivations that we obtained by
reviewing the steps of the HGA mentioned above.
\begin{itemize}
\item From the mechanism of the crossover and mutation operations,
we can find that the algorithm is self-adapted and can maintain the
population's diversity. This characteristic may be utilized in
dynamic environments.

\item The HGA will find the global optimum successfully
only if the points obtained from the algorithm ¡®drop¡¯ into the
neighborhood of the global optimum. Therefore, when the environment
has changed, if the global optimum is varied slightly, the HGA could
still track it successfully because of the function of the local
search.
\end{itemize}

Based on these motivations, we will make a slight revision of the
HGA and apply this HGA in some dynamic environments. The details are
mentioned in the next Section.

\section{Dynamic environments}
The dynamic optimization problem can be described as:
\begin{equation}
\min\:f(\boldsymbol{x},t)\;\;\boldsymbol{x}\in S,\,t\in T
\end{equation}
The problem depends on both $\boldsymbol{x}$ and an additional
parameter $t\in T$ (the {\em time}). Generally, the objective
function might be different after each function evaluation. In this
paper, we assume that the functional change is observable and the
objective function remains constant within specific time intervals
$[t_k ,t_k + \Delta t_k )\,(k=0,1,2,\ldots)$. Moreover, the dynamics
of the objective function and the dynamics of the genetic algorithm
are synchronized by identifying $t$ with the number of evaluations
of the algorithm and by keeping $f$ constant. Furthermore, $\Delta
t_k=:\Delta g$ ($\Delta g$ denotes the number of the evaluations
between the time intervals) is also assumed to be constant, such
that the objective function changes every $\Delta g$ evaluations in
case of a genetic algorithm.

An often used benchmark problem in some literature to test GA in
dynamic environments (such as \cite{back1998446}) is
\[f(\boldsymbol{x},t)=\sum^n_{i=1}(x_i-\delta_i(t))^2.\]
This benchmark is too simple for the HGA described in this paper.
Instead, three dynamical environments are derived from the model:
\begin{equation}\label{stationary}
f(\boldsymbol{x}) = 10n + \sum\limits_{i = 1}^n {[x_i^2  - 10\cos
(2\pi x_i )]},
\end{equation}

The figure of \eqref{stationary} when $n=2$ on
$[-5.12,\,5.12]\times[-5.12,\,5.12]$ is shown in Figure 1. The
global optimum is $(0, 0,\ldots, 0)^T$ and the optimum value is 0.

\begin{figure}
\caption[ ]{The figure of \eqref{stationary} when $n=2$ on
$[-5.12,\,5.12]\times[-5.12,\,5.12]$} \centering
\includegraphics[width=8cm]{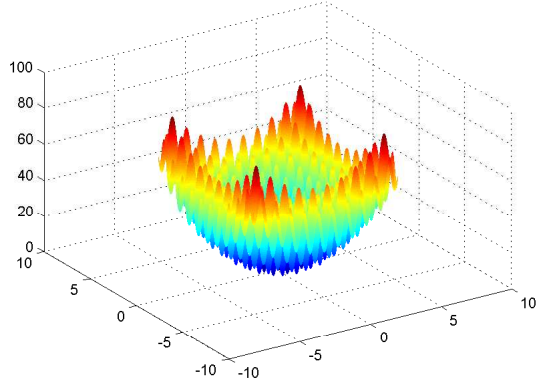}
\end{figure}

The dynamical environments are generated by translating the base
function along three different types of dynamics (linear, circular,
and random) based on

\begin{equation}\label{benchmark}
f({\boldsymbol{x}},\,t) = 10n + \sum\limits_{i = 1}^n {[(x_i  -
\delta _i (t))^2  - 10\cos (2\pi (x_i  - \delta _i (t))]}
\end{equation}

where $t \in \mathds{N} $ denotes the time counter;
$\boldsymbol{\delta} (t) = (\delta _1 (t), \ldots ,\delta _n (t))$
denotes the dynamical offset. The latter depends on three
parameters, namely the {\em update type} (linear, circular, and
random), the {\em update frequency} $\Delta g$ (i.e., the number of
evaluations between each update of offset vector), and the {\em
severity} $s$ (i.e., a factor that determines the amount of the
displacement per object variable).

The three different types of dynamics are defined as follows:
\begin{itemize}
\item {\em Linear dynamics}: $\delta _i (0) = 0,\,i \in \{ 1, \ldots ,n\} $
\begin{equation}\label{linear}
\delta _i (t + 1) = \left\{ \begin{array}{l}
 \delta _i (t) + s,\quad (t + 1)\bmod \,\Delta g = 0 \\
 \delta _i (t),\quad \quad \,{\rm{else}} \\
 \end{array} \right.
\end{equation}
\item {\em Circular dynamics}:
\begin{equation}\label{eq9}
\delta _i (0) = \left\{ \begin{array}{l}
 0,\quad i\;{\rm{odd}} \\
 s,\quad i\,\,{\rm{even}} \\
 \end{array} \right.
\end{equation}
\begin{equation}\label{circular}
\delta _i (t + 1) = \left\{ \begin{array}{l}
 \delta _i (t) + s \cdot c(i,t),\quad (t + 1)\bmod \,\Delta g = 0 \\
 \delta _i (t),\quad \quad \,{\rm{else}} \\
 \end{array} \right.
\end{equation}
where
\begin{equation}
c(i,t) = \left\{ \begin{array}{l}
 \sin \left( {\frac{{2\pi \left\lfloor {t/\Delta g} \right\rfloor }}{\gamma }} \right),\quad i\;{\rm{odd}} \\
 \cos \left( {\frac{{2\pi \left\lfloor {t/\Delta g} \right\rfloor }}{\gamma }} \right),\quad i\,\,{\rm{even}} \\
 \end{array} \right.
\end{equation}
Here, the expression $\left\lfloor {t/\Delta g} \right\rfloor $
 is the number of applications of the dynamics
within evaluations $\{ 0,1, \ldots ,t\} $ and $\gamma $
 represents the number of applications of
the dynamics required to cycle the values of the offset vector.
Following \cite{back1998446}, $\gamma  = 25$ is used in the
experiments performed here.
\item {\em Random dynamics}: $\delta _i (0) = 0\,\forall i \in \{ 1, \ldots ,n\} $
\begin{equation}\label{random}
\delta _i (t + 1) = \left\{ \begin{array}{l}
 \delta _i (t) + s \cdot N_i (0,1),\quad (t + 1)\bmod \,\Delta g = 0 \\
 \delta _i (t),\quad \quad \,{\rm{else}} \\
 \end{array} \right.
\end{equation}
$N_i (0,1)$ denotes a normally distributed random variable with
expectation zero and variance one.
\end{itemize}

In the three cases mentioned above, the update frequency $\Delta g$
determines whether an update of $\boldsymbol{\delta}$ will happen or
not. The parameter $s$ determines the severity of the update in each
dimension.

In order to let the HGA more suitable for dynamic environments, a
slight revision is made. When a change is detected (notice that we
have assumed the change is observable), in \eqref{mutation} we
temporarily replace $\pi$ as a random number and $Y_1-Y_2$ a random
vector in one generation to make the individuals scatter on the
solution set, then restore them.

\section{Experimental Results}
We used the HGA on the benchmark mentioned in Section 3 and compared
the results with some algorithms in the literature. The first is a
variation of standard genetic algorithm with a memory module
(\cite{branke19991875}). It is denoted as {\bf SGA/M}. Another
algorithm introduces a number of random immigrants when a change
occurs \cite{grefenstette1992137}. In this paper, $25$ random
immigrants are migrated into the current population, as suggested in
\cite{branke2001}. This implementation of random immigrants is
accompanied by a memory module to enhance its performance. It will
be referred to as {\bf RI25/M}. The third algorithm we used is the
hyper-mutation algorithm \cite{cobb1993523} in which a memory module
is also included. This implementation will be referred to as {\bf
HM/M}. The last algorithm, referred to as {\bf ERS}, is proposed in
\cite{garrett2002359}. It unified genetic algorithm and a hill-climb
search in dynamic optimization and can be regarded as another HGA.

In the experiments, each test is run $20$ times. We use the
parameters same as \cite{branke2000299, branke19991875} which are
convenient for comparison. For all algorithms we use a population
size of $100$, a crossover rate of $0.6$, a mutation rate of $0.2$.
In addition, we use {\bf SBX} crossover and mutation with
distribution index of $0.7$. Binary tournament selection is adopted
in recombination and replace schemes. In addition, it is also used
in the selection of individuals for reevaluation. A single elitism
is also used to preserve the performance of the best individual at a
given generation. A maximum fitness evaluations is set to $500\,000$
for all implementations. Each dimension in decision space is bound
between $-50$ and $50$.

There are several performance indices that have been suggested to
measure the performance of dynamic evolutionary algorithms. In this
paper, we use {\em off-line error variation} \cite{dejong1975} as
means to quantify the performance of the proposed algorithm.

Off-line error variation index \cite{dejong1975} is the most
commonly used performance index in GAs in dynamic environments, and
it is obtained as the average of the error between the true optimal
fitness and the best fitness at each evaluation. It is
mathematically expressed as:
\begin{equation}
\overline {e_{{\rm{off - line}}} }  = \frac{1}{T}\sum\limits_{i =
1}^T {(f_{{\rm{true}}}  - f_{{\rm{best}}}^i )}
\end{equation}
where $i$ is the evaluation counter; $T$ is the total number of
evaluations considered; $f_{\rm{true}}$ is the true optimum solution
that is updated whenever a change occurs; and finally
$f^i_{\rm{best}}$ is the best individual out of the evaluations
starting from the most recent occurrence of the change until the
current evaluation.

\begin{table}
\centering \caption{Off-line error variation after $50\,000$
evaluations in different dimensions (Severity $s=0.1$, update
frequency $\Delta g=5\,000$)}
\begin{tabular}{|m{1.5cm}|c|c|c|c|c|c|}
\hline
\centering Moving Type& Dimension & SEA/Mem & RI25/Mem & HM/Mem & ERS & HGA\\
 \hline
 & 5 & 16.25 & 18.92 & 22.51 & 2.85 & {\bf2.10}\\
\cline{2-7}
 & 10 & 61.21 & 65.70 & 87.24 & 39.51 & {\bf4.48}\\
\cline{2-7}
\centering Linear & 50 & 528.32 & 557.37 & 738.44 & 1015.1 & {\bf25.43}\\
\cline{2-7}
 & 100 & 1273.2 & 1422.7 & 1815.4 & 2757.5 & {\bf48.31}\\
\cline{2-7}
 & 200 & 2376.6 & 3349.8 & 3473.7 & 7772.9 & \bf88.90\\
\hline
 & 5 & 26.22 & 23.65 & 25.01 & 3.27 & \bf0.73\\
\cline{2-7}
 & 10 & 107.37 & 96.96 & 103.61 & 32.94 & \bf1.95\\
\cline{2-7}
\centering Circular & 50 & 1034.6 & 931.70 & 1019.2 & 904.96 & \bf20.39\\
\cline{2-7}
 & 100 & 2141.8 & 2135.9 & 2446.4 & 2685.8 & \bf23.07\\
\cline{2-7}
 & 200 & 4096.8 & 4792.2 & 4688.1 & 7183.0 & \bf46.87\\
\hline
 & 5 & 22.10 & 20.42 & 22.18 & 3.65 & \bf2.39\\
\cline{2-7}
 & 10 & 86.17 & 94.61 & 99.41 & 33.52 & \bf5.87\\
\cline{2-7}
\centering Random & 50 & 775.13 & 818.55 & 806.51 & 923.08 & \bf49.98\\
\cline{2-7}
 & 100 & 1478.6 & 1570.6 & 1855.4 & 2574.5 & \bf97.71\\
\cline{2-7}
 & 200 & 3384.7 & 3154.7 & 3676.5 & 6942.4 & \bf236.46\\
\hline

\end{tabular}
\end{table}

\begin{table}
\centering \caption{Off-line error variation after $50\,000$
evaluations in different severities (Dimension $n=15$, update
frequency $\Delta g=5\,000$)}
\begin{tabular}{|m{1.5cm}|c|c|c|c|c|c|}
\hline
\centering Moving Type & Severity & SEA/Mem & RI25/Mem & HM/Mem & ERS & HGA\\
\hline
 & 0.01 & 93.00 & 102.61 & 173.89 & 61.77 & \bf1.63\\
\cline{2-7}
\centering Linear & 0.1 & 98.18 & 114.26 & 177.60 & 85.00 & \bf6.83\\
\cline{2-7}
 & 0.5 & 116.79 & 179.30 & 177.07 & 114.12 & \bf22.48\\
\hline
 & 0.01 & 111.20 & 101.06 & 173.58 & 104.59 & \bf0.09\\
\cline{2-7}
\centering Circular & 0.1 & 194.73 & 191.84 & 215.23 & 99.91 & \bf2.27\\
\cline{2-7}
 & 0.5 & 343.03 & 307.68 & 312.17 & 113.23 & \bf10.94\\
 \hline
 & 0.01 & 112.40 & 137.16 & 176.21 & 81.97 & \bf8.99\\
\cline{2-7}
\centering Random & 0.1 & 177.56 & 173.75 & 202.77 & 88.94 & \bf10.94\\
\cline{2-7}
 & 0.5 & 235.01 & 223.54 & 244.05 & 138.72 & \bf17.08\\
\hline
\end{tabular}
\end{table}

\begin{table}
\centering \caption{Off-line error variation after $50\,000$
evaluations in different update frequencies (Dimension $n=15$,
severity $s=0.1$)}
\begin{tabular}{|m{1.5cm}|m{2cm}|c|c|c|c|c|}
\hline
\centering Moving Type & \centering Update frequency & SEA/Mem & RI25/Mem & HM/Mem & ERS & HGA\\
\hline
 & \centering $1\,000$ & 111.93 & 106.91 & 151.39 & 44.24 & \bf85.15\\
\cline{2-7}
\centering Linear & \centering $2\,500$ & 75.26 & 83.73 & 105.28 & 28.67 & \bf19.29\\
\cline{2-7}
 & \centering $10\,000$ & 72.13 & 56.70 & 88.43 & 25.72 & \bf0.33\\
\hline
 & \centering $1\,000$ & 132.45 & 135.09 & 179.36 & 42.23 & \bf44.40\\
\cline{2-7}
\centering Circular & \centering $2\,500 $& 92.49 & 96.32 & 115.54 & 34.56 & \bf9.81\\
\cline{2-7}
 & \centering $10\,000$ & 73.27 & 84.02 & 85.90 & 33.40 & \bf1.31\\
 \hline
 & \centering $1\,000$ & 111.78 & 119.19 & 186.77 & 43.72 & \bf15.36\\
\cline{2-7}
\centering Random & \centering $2\,500$ & 95.51 & 97.36 & 114.20 & 29.81 & \bf10.69\\
\cline{2-7}
 & \centering $10\,000$ & 86.70 & 86.75 & 92.98 & 26.45 & \bf9.73\\
\hline

\end{tabular}
\end{table}

Firstly, we fix the severity $s$ in \eqref{linear}, \eqref{eq9},
\eqref{circular}, and \eqref{random} as $0.1$, the update frequency
$\Delta g=5\,000$, and test different functional dimensions of
\eqref{benchmark}. Secondly, we set the functional dimension as $15$%
, the same update frequency, and test all algorithms on different
severities ($0.01$, $0.1$, and $0.5$). Finally, we set the
functional dimension as $15$, severity $s=0.1$, and test all
algorithms on different update frequencies. Tables 1, 2, and 3 show
off-line error variations after $50\,000$ evaluations of the HGA and
other algorithms in the dynamic environments with different moving
types (Linear, Circular, and Random). In all these cases, the HGA
can achieve much better performance than other algorithms mentioned
in this paper.

\section{Conclusions}
An HGA proposed in \cite{yuan2008924} has been tested by different
functional dimensions, update frequencies, and displacement
strengths in different types of dynamics. Compared with some other
existing evolutionary algorithms for dynamic environments, the HGA
has been illustrated its better capability to track the dynamic
optimum based on the results. This HGA can be an alternative for
dynamic optimization problems. A more detailed investigation of the
working principles of the HGA and how to apply this HGA to other
kinds of dynamic optimization problems (such as online optimization)
will be the subjects of further work.

\section*{Acknowledgments}
The authors are very grateful to the anonymous referee and Professor
Qing-Long Han for their comments and suggestions that lead to a
significant improvement of the paper.

\vspace{5cm} 

\end{document}